\documentclass[letterpaper, 10 pt, conference]{ieeeconf}  % Comment this line out if you need a4paper

\IEEEoverridecommandlockouts                              % This command is only needed if 
\usepackage{amsmath}
\usepackage{graphicx}
\usepackage{wrapfig}
\usepackage{algpseudocode}
\usepackage[ruled,vlined,linesnumbered]{algorithm2e}
\usepackage{makecell}
\usepackage{booktabs}
\usepackage{siunitx}

\usepackage{setspace}
\usepackage{balance}

\title{\LARGE \bf
Whisker-based Active Tactile Perception for Contour Reconstruction 
}

\author{Yixuan Dang$^{1}$, Qinyang Xu$^{1}$, Yu Zhang$^{1}$, Xiangtong Yao$^{1}$, Liding Zhang$^{1}$, Zhenshan Bing$^{1,2}$, \\ Florian Roehrbein$^{3}$, Alois Knoll$^{1}$% <-this % stops a space
    \thanks{$^{1}$Authors from the Technical University of Munich, Munich, Germany. \{yixuan.dang, qinyang.xu, zy.zhang, xiangtong.yao, liding.zhang, zhenshan.bing, k\}@tum.de}% 
    \thanks{$^{2}$Z. Bing is also with the State Key Laboratory for Novel Software Technology and the School of Science and Technology, Nanjing University (Suzhou Campus), China. \textit{(Corresponding author: Zhenshan Bing.)}}%
    \thanks{$^{3}$Author from the Chemnitz University of Technology, Chemnitz, Germany. florian.roehrbein@informatik.tu-chemnitz.de}%
}

\begin{document}

\setstretch{0.95}

\maketitle
\thispagestyle{empty}
\pagestyle{empty}

\begin{abstract}

Perception using whisker-inspired tactile sensors currently faces a major challenge: the lack of active control in robots based on direct contact information from the whisker. To accurately reconstruct object contours, it is crucial for the whisker sensor to continuously follow and maintain an appropriate relative touch pose on the surface. This is especially important for localization based on tip contact, which has a low tolerance for sharp surfaces and must avoid slipping into tangential contact. In this paper, we first construct a magnetically transduced whisker sensor featuring a compact and robust suspension system composed of three flexible spiral arms. We develop a method that leverages a characterized whisker deflection profile to directly extract the tip contact position using gradient descent, with a Bayesian filter applied to reduce fluctuations. We then propose an active motion control policy to maintain the optimal relative pose of the whisker sensor against the object surface. A B-Spline curve is employed to predict the local surface curvature and determine the sensor orientation. Results demonstrate that our algorithm can effectively track objects and reconstruct contours with sub-millimeter accuracy. Finally, we validate the method in simulations and real-world experiments where a robot arm drives the whisker sensor to follow the surfaces of three different objects.

\end{abstract}
\section{INTRODUCTION}

Many animals use their vibrissae, or whiskers, to navigate and perceive their environment in the dark, confined spaces~\cite{carvell1990biometric, grant2013evolution, bing2023lateral}. This tactile sensing complements optical perception, with the flexibility of whiskers allowing them to maneuver through various scenarios without obstruction. However, in most situations, whisker sensing is employed passively to help animals interact with complex environments. For example, mice leap over obstacles after sensing bumps around the lower areas of their rostral whiskers~\cite{warren2021rapid}. Notably, some rats engage in sophisticated active sensing~\cite{huet2014search, zweifel2021dynamical} by rotating their vibrissae arrays to extract texture features and shapes from their surroundings. This active sensing behavior serves as the primary inspiration for the method we present here.

Whisker-inspired tactile sensing is also advantageous for enhancing robotic perception~\cite{prescott2009whisking, kaneko1998active, xiao2023complacent, huang2024optimizing}. For instance, navigating cluttered and unstructured environments requires robots to be acutely aware of nearby objects in close proximity, where their motion is often restricted and visibility severely limited due to occlusions between objects from optical sensors. In such scenarios, perception through multiple modalities, including touch, becomes crucial~\cite{kemp2007challenges}, and the whisker sensor emerges as an ideal solution, considering its flexibility and lightweight. The reaction force of contact from an object is transmitted into deflection along the whisker shaft and deformation on the root device, or embedded mechanoreceptors~\cite{szwed2006responses}, which minimizes disturbance to any free-standing light objects and allows the robot to maneuver freely in tight spaces. Although this whisking system can accurately reconstruct surfaces by locating contact positions along the shaft, a significant challenge remains: Without active motion control to continuously follow the unknown surface, this reconstruction can only capture partial features of the object, which remains as a passive perception. While several notable studies have focused on applying active perception or tactile servoing to various other sensors, such as TacTip and the iCub fingertip~\cite{7837664, 7466131}, few have attempted to implement active control based on stimuli from whisker sensors. This gap serves as the primary focus of this paper.

\begin{figure}[t!]
  \centering
%   \framebox{\parbox{3in}{We suggest that you use a text box to insert a graphic (which is ideally a 300 dpi TIFF or EPS file, with all fonts embedded) because, in an document, this method is somewhat more stable than directly inserting a picture.
% }}
  \includegraphics[width=0.98\linewidth]{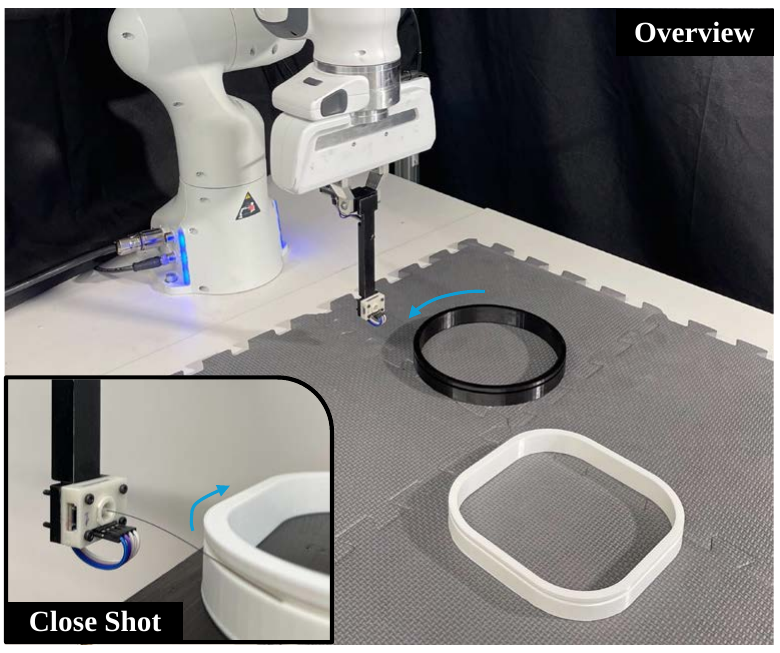}
  \caption{Demonstration of active tactile perception based on whisker-inspired sensor for contour reconstruction. The sensor is driven by a robot arm to continuously follow and make contact with the unknown object's surface.} 
  \label{figurelabel}
  \vspace{-1.5em}
\end{figure}

Some previous research relies on tip contact of the whisker shaft to locate objects~\cite{9817376, 9114501}. While this approach causes even less disturbance on objects with light contact force, it is prone to detachment from the object or slipping into tangential contact, leading to a limited measurement range and a low tolerance for sharp surfaces. Although tangential contact experiences less significant fluctuations, its trajectory is still strongly influenced by friction, texture, surface defects, and local curvature~\cite{merker2020effects}. Therefore, it becomes even more critical for the whisker sensor to continuously maintain an optimal contact pose against the object to ensure steady and accurate contact localization.

In this paper, we first develop a root suspension system with three integrated spiral arms to create a magnetically transduced whisker sensor that is more compact and robust. To test our active control strategy, we propose a direct method for extracting the 2D coordinates of the tip contact position based on the characterized deflection profile of the whisker. This profile includes the known total length of the whisker shaft, a root position with rotation and linear translation at the center of the suspension device, and a calibrated whisker model built by sampling deflection measurements at various contact positions. The tip position is calculated using gradient descent, with fluctuations further reduced by applying a Bayesian filter, assuming a constant contact state. We then develop a combination of robot control and whisker-based tactile perception within a single loop. To maintain the sensor's perceived orientation relative to the edge, a B-Spline curve is used to predict the local surface curvature, with a fixed total linear displacement between iterations. The sensor is driven to move tangentially while adaptive adjustments are made to its normal displacement based on the deflection magnitude and a PID controller.

In short, the key contributions are as follows: 
\begin{itemize} 
\item We construct a magnetically transduced whisker sensor based on a root suspension system with three integrated spiral arms, resulting in a more compact design (with a radius of only \SI{3.36}{mm}) and improved robustness. 
\item We propose a direct method based on gradient descent to extract the 2D tip contact position with reduced fluctuations, achieving sub-millimeter accuracy (with an average distance of \SI{0.08}{mm}) and ease of computation.
\item We develop a combination of active motion control and tactile perception within a single loop, using a B-Spline curve to enable the sensor to continuously follow the object's edge and reconstruct its entire surface. 
\end{itemize}

\section{Background and Related Work}

\subsection{Whisker-inspired Tactile Sensor}

Various studies have explored different structural designs for whisker sensors, with the key differences lying in their basic transduction principles—specifically, how the contact force from the whisker is transmitted into stimulus signals at the base. One of the most important designs is the magnetically transduced whisker~\cite{9830882, 9982122}, which uses a Hall effect sensor placed beneath an axially magnetized magnet and suspension system. This design is accurate, compact, and offers high angular deflection resolution, although this structure may suffer from fabrication tolerances~\cite{8968518}. Another popular method is the MEMS barometer-based whisker~\cite{9813357, 10610850}, where the whisker shaft is directly attached to the receptor surface and connected to barometers and a PCB layout. This approach is robust and easily integrated with other whiskers to form a sensory array. Additionally, 6-axis force/torque sensors are widely used as transducers~\cite{huet2017tactile, collinson2021tapered}, providing accurate estimations of tangential contact positions based on the acquired bending moment. However, they are often too bulky and expensive for practical use in robotics. Optical-based approaches face similar challenges~\cite{9366394, 10160408}, requiring substantial mounting space and offering limited sensitivity.

Whisker sensors have proven useful to reconstruct the object contours~\cite{9982122, 9817376, merker2021vibrissa}. It relies on an accurate and robust contact localization, which however is extremely challenging for two main reasons: 1) Different contact states, varying reaction forces, and texture features can cause friction or slip, leading to unpredictable fluctuations in estimation; 2) Differentiating the actual position of a tangential contact, as shown in Fig.2B, is difficult because there is no direct, unique mapping from the single deflection measurement at the whisker root to a specific point along the shaft. Generally, tip contact causes more fluctuation than tangential contact due to friction, yet it is more common unless the whisker is overly deflected or contacts an object mid-shaft. Previous studies have also proposed solutions for tangential contact estimation, but these often require torque measurement~\cite{huet2017tactile, collinson2021tapered} with complex models to fit unique mapping or extra proprioceptive sensing~\cite{9982122}.

\begin{figure}
  \centering
%   \framebox{\parbox{3in}{We suggest that you use a text box to insert a graphic (which is ideally a 300 dpi TIFF or EPS file, with all fonts embedded) because, in an document, this method is somewhat more stable than directly inserting a picture.
% }}
  \includegraphics[width=\linewidth]{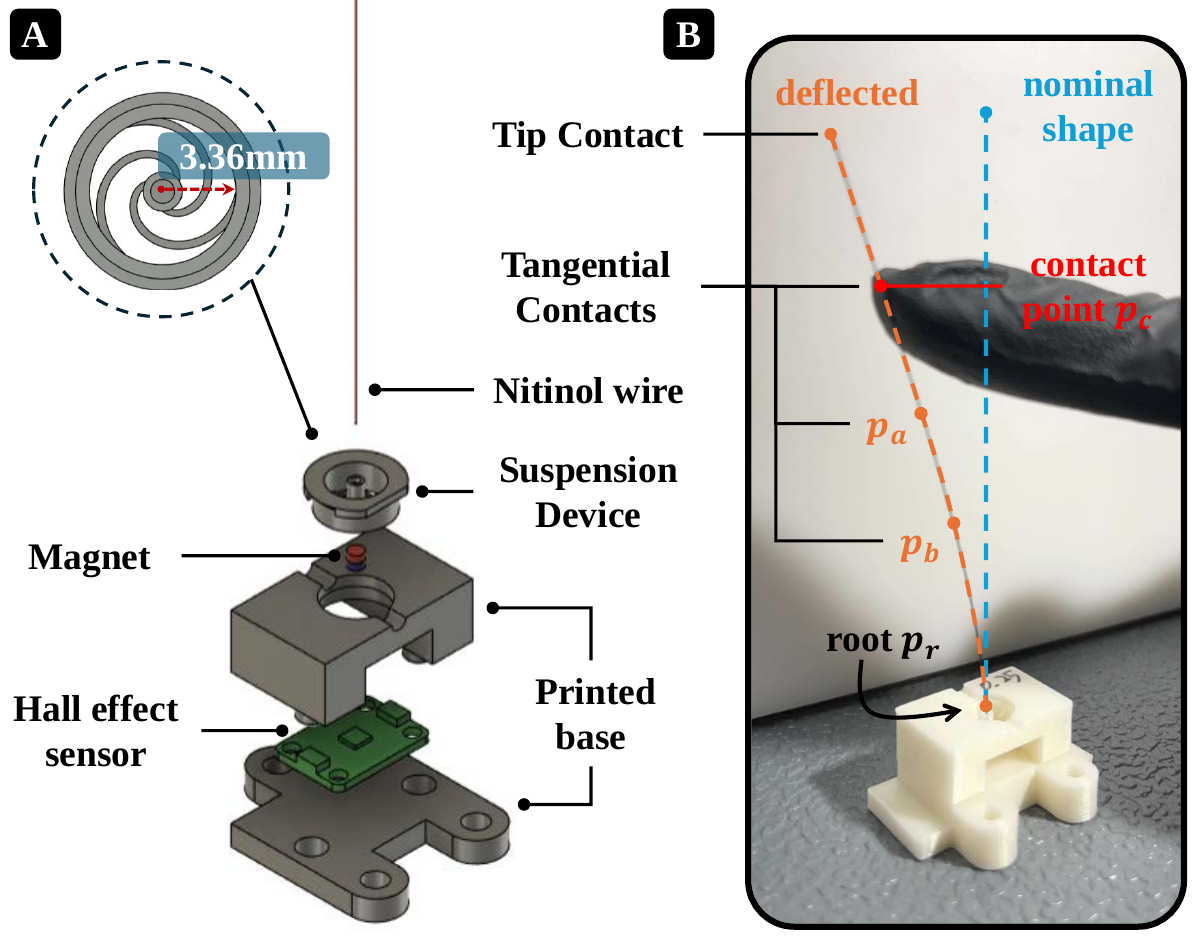}
  \caption{A) The basic structure of our whisker-inspired tactile sensor is based on a magnetically transduced principle. B) The contact force on $p_c$ leads to a deflection along the whisker shaft, deformation on the suspension device, and rotation on the magnet, and ends up with a magnetic flux change measurement on the Hall effect sensor. The tangential contacts at $p_a$, $p_b$, and $p_c$ result in the same deflection measurement on the root.} 
  \label{figurelabel}
  \vspace{-1.0em}
\end{figure}

\subsection{Active Tactile Perception}

% \begin{figure*}
%   \centering
% %   \framebox{\parbox{3in}{We suggest that you use a text box to insert a graphic (which is ideally a 300 dpi TIFF or EPS file, with all fonts embedded) because, in an document, this method is somewhat more stable than directly inserting a picture.
% % }}
%   \includegraphics[width=\linewidth]{figure/Fig3.pdf}
%   \caption{A) Setup for motorized stage to calibrate the whisker sensor. B) Default root position $\boldsymbol{p}_r$ and extracted general deflection profile on whisker sensor, maps tangential contact point $(x^b_t, y^b_t)$ to measured magnetic flux change $z$ on Hall effect sensor. C) Extract tip position by track the profile from tangent direction and loss direction based on gradient descent.}
%   \label{figurelabel}
%   \vspace{-1.0em}
% \end{figure*}

\begin{figure*}
  \centering
%   \framebox{\parbox{3in}{We suggest that you use a text box to insert a graphic (which is ideally a 300 dpi TIFF or EPS file, with all fonts embedded) because, in an document, this method is somewhat more stable than directly inserting a picture.
% }}
  \includegraphics[width=\textwidth]{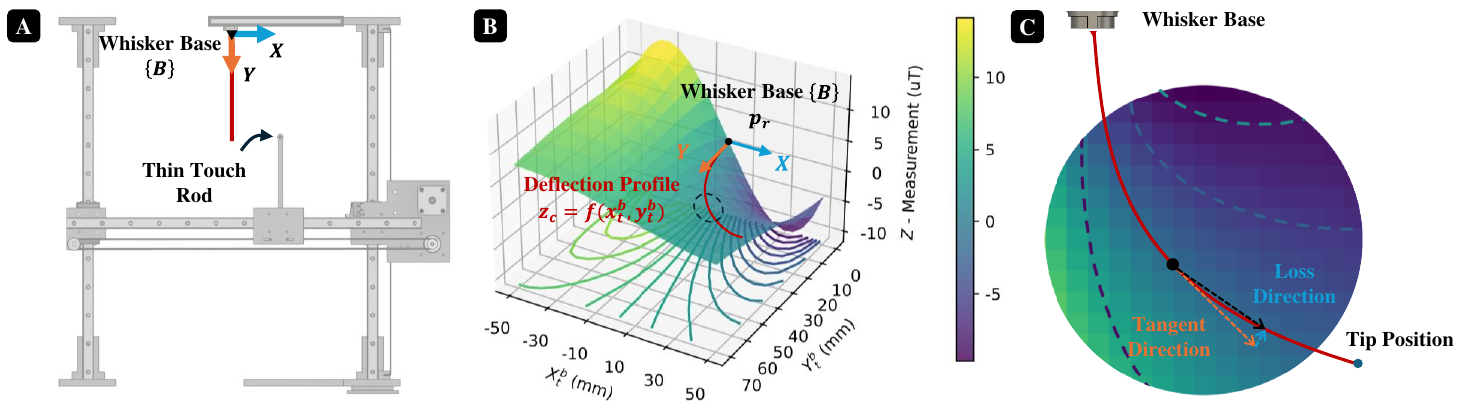}
  \caption{A) Setup for the 2-dimensional motorized stage to calibrate the whisker sensor. B) Default root position $\boldsymbol{p}_r$, extracted general deflection profile on whisker sensor and surface which maps tangential contact point $(x^b_t, y^b_t)$ to measured magnetic flux change on Hall effector sensor $z$. C) Extract tip position by tracking the profile from the tangent direction and loss direction based on gradient descent.}
  \label{figurelabel}
  \vspace{-1.0em}
\end{figure*}

Many previous studies have focused on developing various tactile sensors, enabling robots to navigate, perform complex manipulations~\cite{10610224}, interact with environments~\cite{10473086}, and notably, actively reconstruct these environments without vision. For instance, Lepora et al.~\cite{7837664, 7466131} proposed a biomimetic active touch approach that integrates perception and tactile servoing into a single control loop, demonstrating the ability to track edges using tactile stimuli from the iCub fingertip and TacTip. Although numerous other exploration strategies have emerged recently, a significant challenge remains: direct contact could cause disturbance in an object's original state, often leading to intermittent tap contacts for sensing. By interpreting it as a SLAM problem, Suresh et al.~\cite{9562060} were able to infer object shapes while simultaneously maintaining continuous contact and accounting for motion induced by tactile interaction.

Based on its flexibility and lightweight, the Whisker sensor presents as another potential solution to the problem by providing continuous contact perception without disturbing objects; however, few studies have explored this avenue. Xiao et al.~\cite{9813357} applied a MEMS-based whisker array to actively extract object features but did so using a spiral movement with discrete contacts, and the whiskers were only used to detect collisions without actual contact localization. A recent study by Kossas et al.~\cite{10610762} proposed a whisker-based navigation algorithm, but it primarily focused on integrating tactile sensation into the robot's motion decision-making process without constructing actual maps.

\section{Methodology}

% \begin{figure}[t!]
%   \centering
% %   \framebox{\parbox{3in}{We suggest that you use a text box to insert a graphic (which is ideally a 300 dpi TIFF or EPS file, with all fonts embedded) because, in an document, this method is somewhat more stable than directly inserting a picture.
% % }}
%   \includegraphics[width=\textwidth]{figure/Fig3.pdf}
%   \caption{A) Default root position $\boldsymbol{p}_r$ and extracted general deflection profile on whisker sensor, maps tangential contact point $(x^b_t, y^b_t)$ to measured magnetic flux change $z$ on Hall effect sensor. B) Extract tip position by track the profile from tangent direction and loss direction based on gradient descent.} 
%   \label{figurelabel}
%   \vspace{-1.0em}
% \end{figure}

\subsection{Sensing System Integration}
The overall mechanism schematic of our magnetically transduced whisker sensor is illustrated in Fig.2A. A suspension system is first constructed using a spring device composed of three integrated flexible spiral arms, directly fabricated via 3D printing using PLA plastic as filament. A nitinol wire (\SI{0.25}{mm} diameter, \SI{75}{mm} length) is selected as the whisker shaft and attached to the top surface of the suspension device. A neodymium permanent magnet, axially magnetized with its field direction aligned with the wire, is attached at the bottom of the device. This upper assembly is supported by a 3D-printed structure positioned above a magnetic sensor (Mlx90393, Adafruit), which is configured to provide a high resolution of \SI{0.15}{\micro\tesla/LSB} for measuring magnetic flux changes in three different directions. As a result, the contact reaction force from the object surface along the whisker shaft is converted into a motion of the magnet, leading to changes in the magnetic field detected by the Hall effect sensor. The magnetic sensor samples data at \SI{300}{Hz} and is connected to an Arduino Nano via I2C communication. Measurement data are transmitted to a computer at the same rate through USB serial communication using a predefined ROS node on the Nano.

\subsection{Whisker Deflection Profile}

The deflection profile of our whisker sensor consists of two primary elements: the root position, denoted as $\boldsymbol{p}_r$, and the measurement model, which maps the tangential contact location $\boldsymbol{c}_t$ to the corresponding deflection measurement $z$, as shown in Fig. 3B. Here, $\boldsymbol{c}_t = (x^b_t, y^b_t)$ represents the position vector in a 2D plane, with the superscript $b$ indicating the reference frame of the whisker sensor base. The deflection magnitude on the shaft is proportional to the bending rotation of the suspension system. We extract the flux change along the Y-axis of the magnetic sensor, which shows the most significant variation, to represent the deflection measurement $z$. Using a calibrated measurement model, a two-dimensional function $z_c = f(x^b, y^b)$ can be formulated to describe the deflection arc shape, where $z_c$ is the corresponding constant measurement value at the given moment.

A 2-dimensional motorized stage is constructed for calibration, as shown in Fig. 3A. This stage consists of two orthogonal joints, with a touch rod attached vertically to the end-effector. The stage is driven by two NEMA17 stepper motors, and the end-effector's motion is precisely measured using two rotary encoders (AS5600, AITRIP). Magnetic measurement data and tangential contact positions are recorded as the motors move in a grid-like pattern, with a fixed step of 3mm along both axes to touch the shaft. Data collection is limited to one side of the whisker sensor’s forward direction, ensuring a more accurate fit of our 5th-order bivariate polynomial model to the calibrated data. A total of 180 data sets are collected. Additionally, the origin of the whisker base frame is set as the default root position, $\boldsymbol{p}_r=(0, 0)$, though this does not precisely match the real situation. A linear displacement may occur at the center of rotation due to the current spring system design, which leaves a gap for future improvement.

\subsection{Tip Contact Localization}

`\textbf{General Principle}: Given the default root position, the deflection profile, and a specified arc distance $L$ (the actual whisker shaft length), the tip position can be extracted by tracing from the start point along the known direction until the end of the trajectory.

We employ a constrained gradient descent method to trace the deflection profile. A loss direction is defined by minimizing the squared distance error from the current deflection measurement to the target to maintain the constant measurement $z_c$. This ensures that the 3-dimensional measurement model from the previous calibration is constrained by the level set $f(x^b, y^b) = z_c$. To prevent the loss direction from merely guiding the trace onto the current deflection profile and getting stuck at a local minimum, a tangent direction is introduced to guide the tracing forward. Finally, by combining the loss and tangent directions, as illustrated in Fig. 3C, the trace advances in fixed step sizes (1e-3 mm), ensuring continuous progress along the deflection profile until the required total movement distance $L$ is completed.

To further reduce computation time, we calculate and resample 20 sets of tip position data to build a characterized model for use in real experiments. This approach enables us to determine the tip contact position within \SI{1}{ms}. Polynomial regression of the calibrated measurement model yields a Root-Mean Squared Error (RMSE) of 0.231 and an R-squared value of 0.9944. The characterized model for tip position calculation achieves RMSE values of 0.0064 and 0.0088 for the X- and Y-axes, respectively, with an R-squared value close to 1.0.

\subsection{Bayesian Filtering}

Bayesian filtering is employed to further reduce fluctuation in tip contact localization caused by friction. A modified version of the Kalman Filter is implemented based on a \textbf{constant state assumption}: With the proposed active control strategy for the sensor's pose, the whisker shaft is maintained to collide with the immobile object at a fixed tip contact position in specific optimal deflection shape. Consequently, the prior prediction for the current step, $\hat{\boldsymbol{x}}_k^-$ ($\boldsymbol{x}$, state vector to represents the 2D tip position under whisker-base frame), remains unchanged and is equal to the estimate from the last step, $\hat{\boldsymbol{x}}_{k-1}$. The prediction and update process is listed below.
\begin{equation}
\begin{array}{ll}
\mathit{Predict:} & \quad \mathit{Update:} 
\\ \hat{x}_k^- = \hat{x}_{k-1} & \quad K_k = P_k^- (P_k^- + R_k)^{-1} 
\\ P_k^- = P_{k-1} + Q & \quad \hat{x}_k = \hat{x}_k^- + K_k (z_k - \hat{x}_k^-) 
\\ & \quad P_k = (1 - K_k) P_k^-
\end{array}
\end{equation}

The measurement noise $\boldsymbol{R}_k$ is updated empirically for each iteration based on the previous $N$ consecutive estimated tip positions from the characterized model. The extracted points and their covariance are used to represent the uncertainty. The process noise $\boldsymbol{Q}$ is fixed with the covariance of $1e^{-5}\mathit{I}_2$, which is found to work well, representing the confidence from active control. As a result, the standard deviation of the filtered results is reduced to 0.078 on the X-axis and 0.033 on the Y-axis.

\subsection{Active Motion Control}

This active control policy makes an action that combines a rotary transformation of the whisker sensor toward the object surface according to a prediction of its local feature based on the B-Spline curve, a normal displacement that actively approaches the edge based on current deflection measurement, and a customized PID controller, and a tangential exploratory movement.

\RestyleAlgo{ruled}
\SetKwComment{Comment}{// }{}
\begin{algorithm}[t!]
\small{
    \caption{\small\text{Whisker Active Perception}}\label{alg:three}
    \KwData{deflection measurement $z$, end-effector state $e$}
    \KwResult{control commands (target orientation $\theta$, linear velocity $v_x, v_y$)}
     initialize the reconstructed contacts deque $Q\gets \emptyset$ \;
     initialize the key points deque $Q_k\gets \emptyset$ \;

    \If{$z$ is received}{
        \If{$abs(z) \ge collisionThreshold$}{
            $contacted \gets 1$;
        }
        \If{$contacted$}{
            \Comment{\small{calculate tip position}}
            $Q.push(z2coordinates(z))$\;
            \If{$len(Q) = filterWindow$}{
                $updateMeasureNoise(Q)$\;
                $filter.predict(), filter.update(Q[-1])$\;
                $\boldsymbol{p}_{cur} \gets transform(filter.x, e)$\;
            }
            \If{$contactPoint$ is keypoint}{
                \Comment{\small{build BSpline curve}}
                $Q_k.push(\boldsymbol{p}_{cur}), c \gets bspline(Q_k)$\;
                $\boldsymbol{p}_{next} \gets c.extrapolate(u_{next})$\;
                $orient \gets arctan(slope(\boldsymbol{p}_{cur}, \boldsymbol{p}_{next}))$\;
            }
            $x_{vel} \gets controller.update(z, x_{vel})$\;
            $y_{vel} \gets constraint(x_{vel}, totalVelocity)$\;
        }
    }
    \Return $\left \{ orient, x_{vel}, y_{vel} \right \} $\;
    }

\end{algorithm}

1) \textit{\textbf{Rotary Action:}} The B-spline curve can construct a continuous analytical model from known observation points and estimate the state of the next non-observation points. The curve $S(\aleph)$ is represented as follows:
\begin{equation}
S(\aleph) = \sum_{j=0}^{n-1} c_j B_{j,k,t}(x)
\end{equation}
where $\aleph\in \left \{ x,y \right \}$ represents the 2D coordinates position in world-fixed frame, $c_j$ is the $j$-th spline coefficients and $n$ is the total number of the sampled control points. $B_{j,k,t}$ are B-spline basis functions of degree $k$ and knots $t$. Further details of the B-spline curve are available in~\cite{8481577}.

\begin{figure}[t!]
  \centering
  \vspace{-0.5em}
%   \framebox{\parbox{3in}{We suggest that you use a text box to insert a graphic (which is ideally a 300 dpi TIFF or EPS file, with all fonts embedded) because, in an document, this method is somewhat more stable than directly inserting a picture.
% }}
  \includegraphics[width=\linewidth]{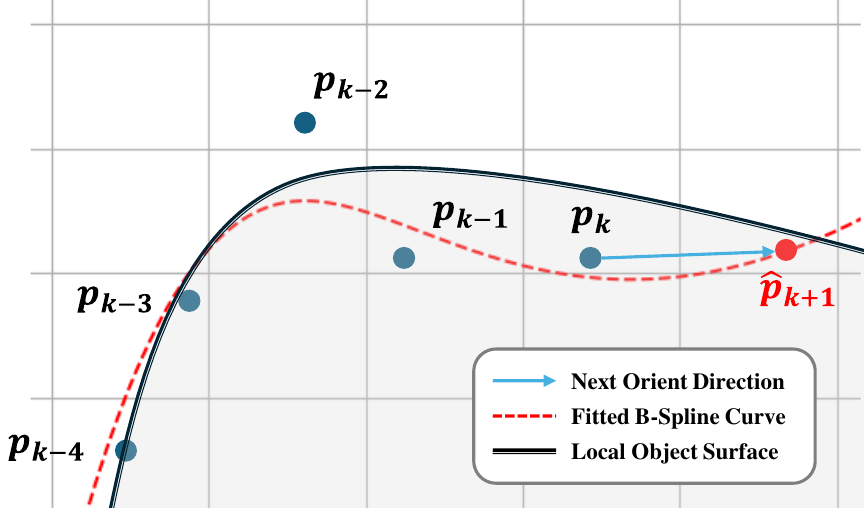}
  \caption{Prediction on next contact point using B-Spline curve and $n$ previous key reconstructs $\left \{ \boldsymbol{p}_{k-i} \mid i=0, 1,...,n-1 \right \}$. The slope between current contact point $\boldsymbol{p}_k$ and next prediction $\hat{\boldsymbol{p}}_{k+1}$ is regarded as the next orient direction.} 
  \label{figurelabel}
  \vspace{-1.5em}
\end{figure}

Suppose the sensor is driven to move along the object contour and leaves reconstructed points based on contacts along its past trajectory; as shown in Fig. 4, the most recent $n$ points are used to construct the B-spline curve. However, the uncertainty from previous direct contact localization may cause fluctuations in the local area, even after filtering. As a result, two consecutive reconstructions might overlap due to noise, leading to significant discontinuity in the B-spline's prediction. Therefore, these key points are extracted every $d$ iteration with an extra fixed interval distance, ensuring that they are orderly distributed and approximately evenly spaced (as the inferences are executed very fast with \SI{300}{Hz}, the contact points will not move far between iterations). In this scenario, the parameterization will be nearly uniform due to the even spacing, allowing for an extrapolated prediction when extending the parameter value beyond the original range as follows:
\begin{equation}
\hat{\boldsymbol{p}}_{k+1} = f(u_{k+1}, \aleph_{k}, \aleph_{k-1}, \aleph_{k-2}, \ldots, \aleph_{k-n+1}) 
\end{equation}
Where $u_{k+1}=1+1/(n-1)$ denotes the parameter value of the next key point, $\left \{ \aleph_{k}, \aleph_{k-1}, \aleph_{k-2}, \ldots, \aleph_{k-n+1} \right \}$ represents the \( n \) selected previous key points. The algorithm is implemented based on SciPy, which by default uses centripetal parameterization. This method helps prevent the clustering of parameter values, leading to better and more stable spline fitting. With the prediction of the next contact position \( \hat{\boldsymbol{p}}_{k+1} \), we can calculate the slope from the current point, $s_{k\to k+1}$ and reconstruct the general direction of the local surface. The action then orients the sensor to the target angle, $arctan(s_{k\to k+1})$ according to the local curvature of the object, functioning as a tactile servoing mechanism.

2) \textit{\textbf{Linear Displacement:}} An optimal deflection target is empirically determined, where fluctuations are minimized, and contact is consistently restricted to the tip of the shaft. After adjusting the orientation in the previous step, an active approaching movement is initiated to move the sensor radially toward the object, ensuring that the deflection magnitude remains at the defined optimal deflection state. The direction of this movement aligns with the sensor's orientation, and its derivative is automatically controlled by a customized proportional-integral-derivative (PID) controller.

A fixed total linear velocity is set to constrain the tangential exploratory movement in relation to the dynamics of the normal displacement. For example, suppose a drastic slope change is detected on the object's surface. In that case, the radial linear velocity will be adjusted accordingly, automatically slowing down the tangential pace along the edge while maintaining the predefined total velocity. This approach also supports the assumption of even spacing from the previous slope calculation. With optimal active control of the end-effector, the sensor should remain parallel to the object surface and maintain a consistent deflection shape, ensuring that a constant total linear displacement of the sensor corresponds to a consistent distance between tip reconstructions across iterations.
\section{Experiments and Results}

In this section, we first compare the localization performance across multiple trials on a flat surface to determine the optimal contact state. The proposed algorithm is then tested in simulations with an open wall trajectory. Finally, real-world tests are conducted on three different objects, and their contours are reconstructed with millimeter-level accuracy.

\begin{figure}[t!]
 % \vspace{-1.0em}
  \centering
%   \framebox{\parbox{3in}{We suggest that you use a text box to insert a graphic (which is ideally a 300 dpi TIFF or EPS file, with all fonts embedded) because, in an document, this method is somewhat more stable than directly inserting a picture.
% }}
  \includegraphics[width=\linewidth]{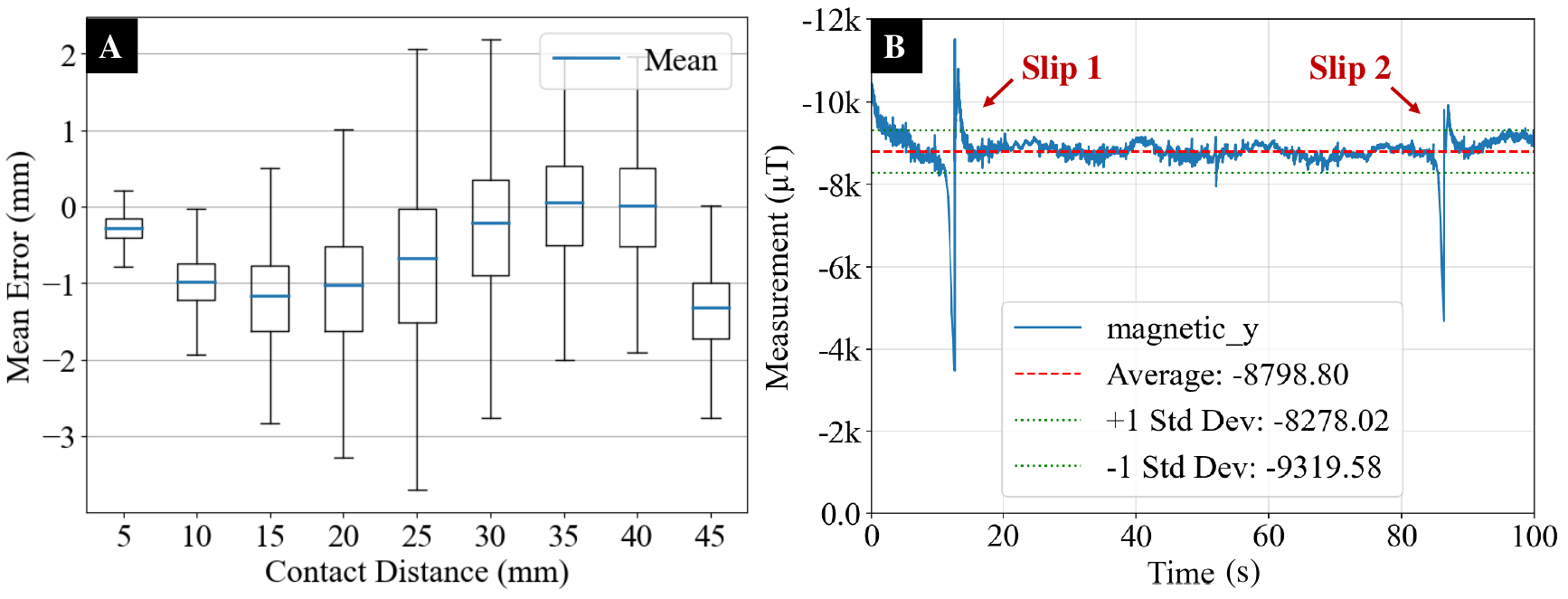}
  \caption{A) Results from 9 trials with varying contact distances are demonstrated. The mean absolute error comes from the minimum distance between contact estimates and ground-truth trajectory, and the corresponding distribution of groups is compared in the plot. B) Magnetic flux change around the Hall effect sensor is recorded during the test. The deflection on the whisker is successfully maintained around its optimal state, which corresponds to a magnetic measurement around \SI{8760}{\micro\tesla}. Two significant slips in the trajectory are also observed from the measurement, which corresponds to the slips in the octagon reconstruction from Fig. 7.}
  \label{figurelabel}
  \vspace{-1.0em}
\end{figure}

\begin{figure}
 \vspace{1.0em}
  \centering
%   \framebox{\parbox{3in}{We suggest that you use a text box to insert a graphic (which is ideally a 300 dpi TIFF or EPS file, with all fonts embedded) because, in an document, this method is somewhat more stable than directly inserting a picture.
% }}
  \includegraphics[width=\linewidth]{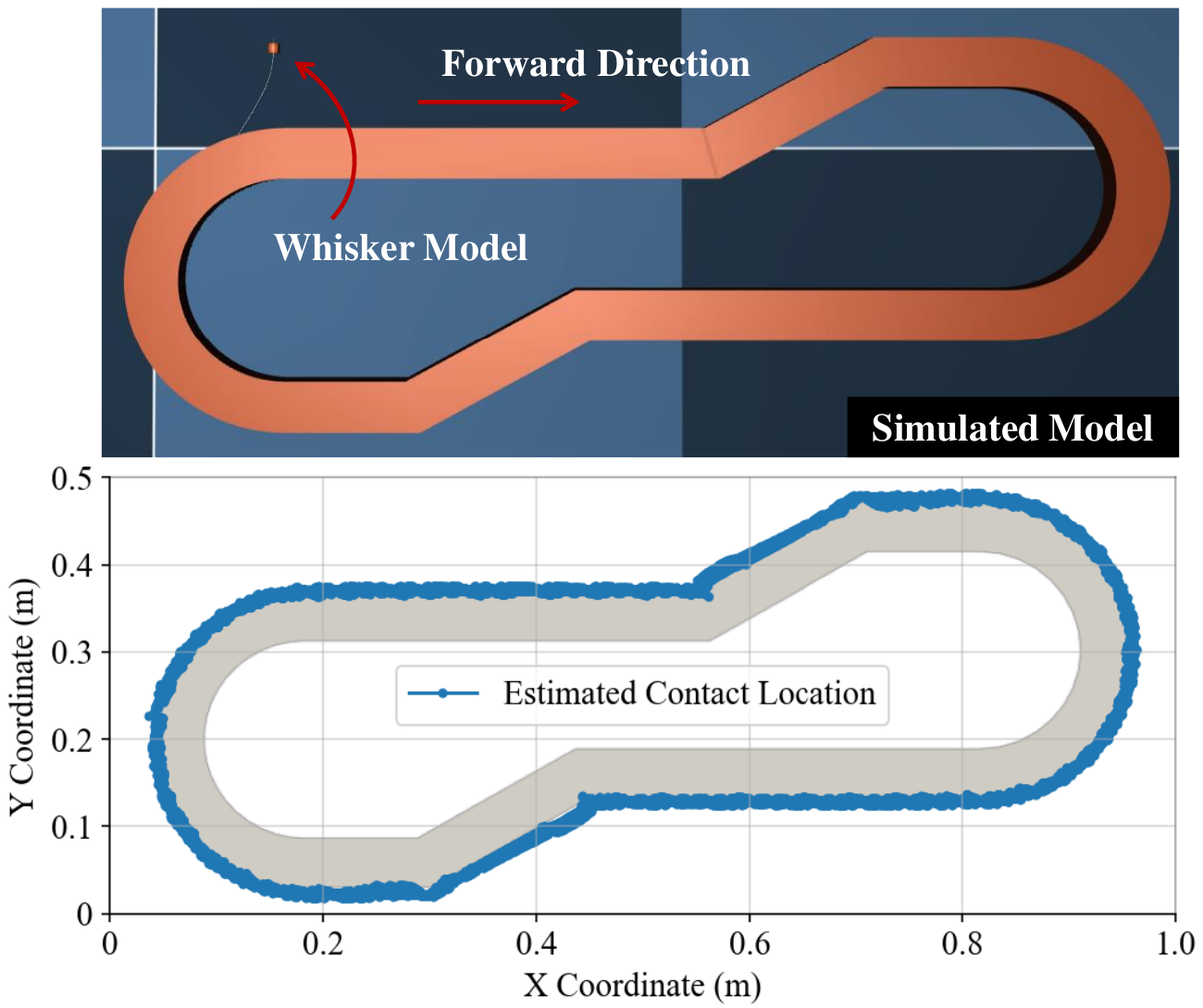}
  \caption{Reconstruction in the simulation environment. A special trajectory is constructed to validate the general effectiveness of the method.}
  \label{figurelabel}
  \vspace{-1.2em}
\end{figure}

\subsection{Optimal Contact State}

The magnet's rotation and the corresponding measurement are proportional to the bending magnitude of the whisker shaft, which directly relates to the contact distance between the sensor and the object surface. To determine an optimal contact deflection target, we first compare the tip localization performance on a flat plane at different contact distances. In this experiment, a printed plane model is attached to the end-effector of our motorized stage. The platform is driven to move at a constant velocity along the X-axis, colliding with the whisker shaft. The experiment consists of 9 trials, with data collected at different distances, each separated by a constant step distance of \SI{5}{mm}. Measurements are sampled at \SI{300}{Hz}, and the total moving distance for each trial is approximately 80mm.

The initial parameter settings for tip localization are consistent across all 9 trials. The algorithm begins calculating the tip position once a collision is detected, indicated by a deflection exceeding the threshold (set by default at \SI{300}{\micro\tesla} above the original static measurement), and the corresponding tip position at this start moment is used as the initial prior mean for the Kalman filter. The initial prior covariance is set to $10.0\mathit{I}_2$, reflecting the uncertainty from the first contact. Given a calculation duration of less than \SI{1}{ms}, the inference is executed at the sensor sampling rate of \SI{300}{Hz}.

Results from trials are compared in Fig. 5A. In the experiments, our proposed localization method tracks the ground-truth contacts with a Euclidean distance error of less than \SI{2}{mm} across all nine trials. Notably, the trial at a height of \SI{5}{mm} achieves the most stable reconstruction, with a minimum standard deviation of 0.25. The trial at a height of \SI{40}{mm} yields the lowest average error of \SI{0.08}{mm}. These results demonstrate that our proposed localization method is capable of reconstructing tip contacts with relatively high accuracy, with the contact distances of \SI{5}{mm}, \SI{30}{mm}, \SI{35}{mm}, and \SI{40}{mm} tracking with sub-millimeter accuracy on average.

\begin{figure}[t!]
  \centering
%   \framebox{\parbox{3in}{We suggest that you use a text box to insert a graphic (which is ideally a 300 dpi TIFF or EPS file, with all fonts embedded) because, in an document, this method is somewhat more stable than directly inserting a picture.
% }}
  \includegraphics[width=\linewidth]{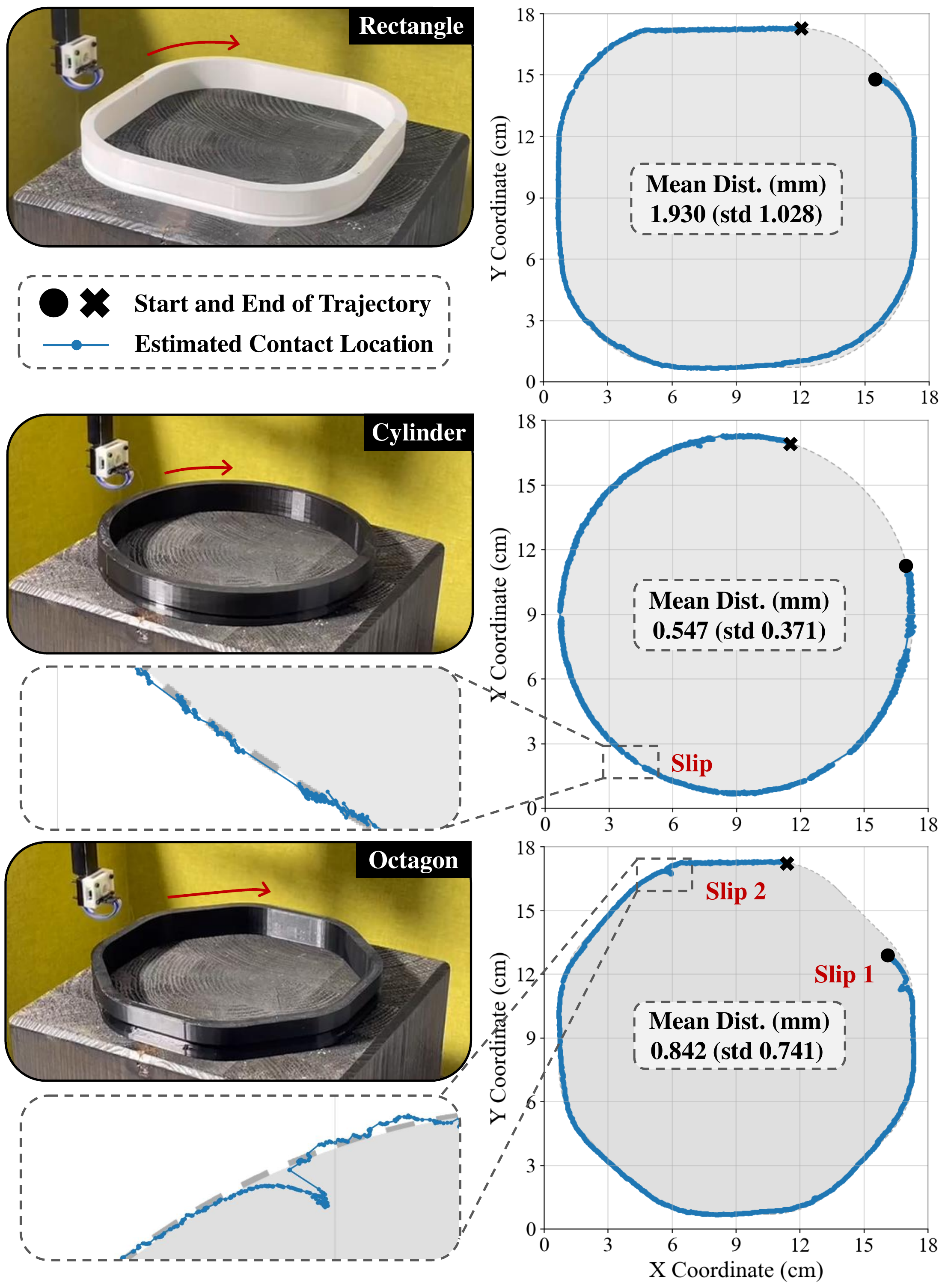}
  \caption{Tracking the contact position and reconstructing the contours when making contact with different objects in the real world. Exemplary slips during the sweep and detailed numerical evaluation on reconstruction are given.}
  \label{figurelabel}
  \vspace{-1.2em}
\end{figure}

Finally, we select the bending state at \SI{40}{mm} as our primary deflection target. It provides the most accurate reconstruction on average, with relatively low variance. Importantly, it strikes a balance: it is neither too close to detaching from the object surface, as observed at \SI{5}{mm}, nor prone to slipping into tangential contact, as seen near \SI{45}{mm}. This choice leaves sufficient space for active control and increases the system's tolerance. The results from the first eight trials show a smooth change in estimates but a drastic decline near \SI{45}{mm}, followed by significant failure in subsequent trials, indicating that it had slipped into tangential contact.

\subsection{Reconstruction with Active Control}

To validate the general effectiveness of our active tactile perception method, we first test it in a simulation environment. The simulated trajectory includes a flat surface, edges with gradually changing curvature, and sharp corners. The whisker body is modeled by connecting 40 elasticity cables in MuJoCo. As shown in Fig. 6, the reconstruction produced is promising. The sensor model can continuously track the surface and navigate around the object without interruption.

Finally, to evaluate the performance further, we collect data from contacts between the whisker sensor and three different objects in the real world. The whisker sensor is mounted directly onto the end-effector of a robot arm (Franka Emika Panda), sampling data at \SI{30}{Hz}. The measurements are transmitted via a ROS-based serial node to the connected computer at the same frequency. The sensor starts to move at a fixed linear speed (\SI{0.01}{m/s}) and is brought into contact with an object by the robot arm from a random position. The calculation on active perception commences immediately after the collision is detected. The scattered contour is reconstructed in real-time on a 2D plane, and active control commands are generated accordingly. The robot arm is simultaneously commanded to drive the sensor roaming around unknown objects using Cartesian velocity control on the end joint, continuing until the maximum rotary range of the joint is reached.

We select three different objects to test our proposed algorithm: a cylinder (diameter: \SI{160}{mm}), a rectangular prism with rounded corners (width: \SI{160}{mm}, radius: \SI{40}{mm}), and an octagonal prism (side length: \SI{70}{mm}, radius: \SI{30}{mm}). Knowing the basic shape and object location allows us to compare the estimated contact points on the object's surface with the ground truth. The tracking results are shown in Fig. 7. Due to the rotary limit on the robot arm's end joint, we could not reconstruct the entire contour in one trajectory, resulting in a small section being missed. However, the available data is sufficient to demonstrate the effectiveness of our methods. 

% Finally, average distance for each objects are: \SI{0.547}{mm} $\pm$ 0.371 for cylinder, \SI{1.930}{mm} $\pm$ 1.028 for rectangle, \SI{0.842}{mm} $\pm$ 0.741 for Octagon. 

Results show that the object surfaces are accurately reconstructed from the contacts, producing distinguishable contour features on three unknown objects. Additionally, Fig. 5B illustrates the corresponding magnetic flux change in the selected direction of the magnetic sensor during the trial on the octagon. The algorithm can recover from a significant slip that occasionally happens in the trajectory. This further confirms that the system is capable of maintaining contact at our predefined optimal deflection state throughout the trajectory. The average bending rotation of the whisker sensor is maintained at \SI{-8798.80}{\micro\tesla}, slightly surpassing the target of \SI{-8760}{\micro\tesla}, which is reasonable since the sensor tends to detach from a convex surface. 
\vspace{-0.2em}
\section{Conclusion}
In this paper, we demonstrate the potential for a reliable active tactile perception using a whisker-inspired sensor. The suspension device is further compacted for our magnetically transduced whisker sensor by introducing three integrated spiral arms. By integrating a tip contact localization with an active motion control policy, the sensor is enabled to continuously follow the unknown surface with its optimal deflection state and reconstruct the contacts with sub-millimeter accuracy. In the future, we aim to further improve our method by: 1) incorporating tangential contact localization into the system to enhance the robustness of active tactile reconstruction on surfaces with varying curvatures, where axial slipping may occur; 2) integrating multiple whiskers into a sensory array and develop an optimization algorithm based on their morphological patterns to improve reconstruction accuracy.

% 3) extend active tactile perception to more complex environments by developing a search and retrieval strategy of whisker sensor on objects.

%                                   % the textheight of the last page by a suitable amount.
%                                   % This command does not take effect until the next page
%                                   % so it should come on the page before the last. Make
%                                   % sure that you do not shorten the textheight too much.

% \balance
\bibliographystyle{IEEEtran}
\bibliography{references}

\begin{thebibliography}{10}
\providecommand{\url}[1]{#1}
\csname url@samestyle\endcsname
\providecommand{\newblock}{\relax}
\providecommand{\bibinfo}[2]{#2}
\providecommand{\BIBentrySTDinterwordspacing}{\spaceskip=0pt\relax}
\providecommand{\BIBentryALTinterwordstretchfactor}{4}
\providecommand{\BIBentryALTinterwordspacing}{\spaceskip=\fontdimen2\font plus
\BIBentryALTinterwordstretchfactor\fontdimen3\font minus
  \fontdimen4\font\relax}
\providecommand{\BIBforeignlanguage}[2]{{%
\expandafter\ifx\csname l@#1\endcsname\relax
\typeout{** WARNING: IEEEtran.bst: No hyphenation pattern has been}%
\typeout{** loaded for the language `#1'. Using the pattern for}%
\typeout{** the default language instead.}%
\else
\language=\csname l@#1\endcsname
\fi
#2}}
\providecommand{\BIBdecl}{\relax}
\BIBdecl

\bibitem{carvell1990biometric}
G.~E. Carvell and D.~J. Simons, ``Biometric analyses of vibrissal tactile
  discrimination in the rat,'' \emph{Journal of Neuroscience}, vol.~10, no.~8,
  pp. 2638--2648, 1990.

\bibitem{grant2013evolution}
R.~A. Grant, S.~Haidarliu, N.~J. Kennerley, and T.~J. Prescott, ``The evolution
  of active vibrissal sensing in mammals: evidence from vibrissal musculature
  and function in the marsupial opossum monodelphis domestica,'' \emph{Journal
  of Experimental Biology}, vol. 216, no.~18, pp. 3483--3494, 2013.

\bibitem{bing2023lateral}
Z.~Bing, A.~Rohregger, F.~Walter, Y.~Huang, P.~Lucas, F.~O. Morin, K.~Huang,
  and A.~Knoll, ``Lateral flexion of a compliant spine improves motor
  performance in a bioinspired mouse robot,'' \emph{Science Robotics}, vol.~8,
  no.~85, p. eadg7165, 2023.

\bibitem{warren2021rapid}
R.~A. Warren, Q.~Zhang, J.~R. Hoffman, E.~Y. Li, Y.~K. Hong, R.~M. Bruno, and
  N.~B. Sawtell, ``A rapid whisker-based decision underlying skilled locomotion
  in mice,'' \emph{Elife}, vol.~10, p. e63596, 2021.

\bibitem{huet2014search}
L.~A. Huet and M.~J. Hartmann, ``The search space of the rat during whisking
  behavior,'' \emph{Journal of Experimental Biology}, vol. 217, no.~18, pp.
  3365--3376, 2014.

\bibitem{zweifel2021dynamical}
N.~O. Zweifel, N.~E. Bush, I.~Abraham, T.~D. Murphey, and M.~J. Hartmann, ``A
  dynamical model for generating synthetic data to quantify active tactile
  sensing behavior in the rat,'' \emph{Proceedings of the National Academy of
  Sciences}, vol. 118, no.~27, p. e2011905118, 2021.

\bibitem{prescott2009whisking}
T.~J. Prescott, M.~J. Pearson, B.~Mitchinson, J.~C.~W. Sullivan, and A.~G.
  Pipe, ``Whisking with robots,'' \emph{IEEE robotics \& automation magazine},
  vol.~16, no.~3, pp. 42--50, 2009.

\bibitem{kaneko1998active}
M.~Kaneko, N.~Kanayama, and T.~Tsuji, ``Active antenna for contact sensing,''
  \emph{IEEE Transactions on robotics and automation}, vol.~14, no.~2, pp.
  278--291, 1998.

\bibitem{xiao2023complacent}
C.~Xiao and J.~Wachs, ``Complacent: A compliant whisker manipulator for object
  tactile exploration,'' in \emph{2023 IEEE/RSJ International Conference on
  Intelligent Robots and Systems (IROS)}.\hskip 1em plus 0.5em minus
  0.4em\relax IEEE, 2023, pp. 10\,184--10\,190.

\bibitem{huang2024optimizing}
Y.~Huang, Z.~Bing, Z.~Zhang, G.~Zhuang, K.~Huang, and A.~Knoll, ``Optimizing
  dynamic balance in a rat robot via the lateral flexion of a soft actuated
  spine,'' in \emph{2024 IEEE International Conference on Robotics and
  Automation (ICRA)}.\hskip 1em plus 0.5em minus 0.4em\relax IEEE, 2024, pp.
  3442--3448.

\bibitem{kemp2007challenges}
C.~C. Kemp, A.~Edsinger, and E.~Torres-Jara, ``Challenges for robot
  manipulation in human environments [grand challenges of robotics],''
  \emph{IEEE Robotics \& Automation Magazine}, vol.~14, no.~1, pp. 20--29,
  2007.

\bibitem{szwed2006responses}
M.~Szwed, K.~Bagdasarian, B.~Blumenfeld, O.~Barak, D.~Derdikman, and
  E.~Ahissar, ``Responses of trigeminal ganglion neurons to the radial distance
  of contact during active vibrissal touch,'' \emph{Journal of
  neurophysiology}, vol.~95, no.~2, pp. 791--802, 2006.

\bibitem{7837664}
N.~F. Lepora, K.~Aquilina, and L.~Cramphorn, ``Exploratory tactile servoing
  with active touch,'' \emph{IEEE Robotics and Automation Letters}, vol.~2,
  no.~2, pp. 1156--1163, 2017.

\bibitem{7466131}
N.~F. Lepora, ``Biomimetic active touch with fingertips and whiskers,''
  \emph{IEEE Transactions on Haptics}, vol.~9, no.~2, pp. 170--183, 2016.

\bibitem{9817376}
Y.~Zhang, S.~Yan, Z.~Wei, X.~Chen, T.~Fukuda, and Q.~Shi, ``A small-scale,
  rat-inspired whisker sensor for the perception of a biomimetic robot: Design,
  fabrication, modeling, and experimental characterization,'' \emph{IEEE
  Robotics \& Automation Magazine}, vol.~29, no.~4, pp. 115--126, 2022.

\bibitem{9114501}
Z.~Wei, Q.~Shi, C.~Li, S.~Yan, G.~Jia, Z.~Zeng, Q.~Huang, and T.~Fukuda,
  ``Development of an mems based biomimetic whisker sensor for tactile
  sensing,'' in \emph{2019 IEEE International Conference on Cyborg and Bionic
  Systems (CBS)}, 2019, pp. 222--227.

\bibitem{merker2020effects}
L.~Merker, S.~J. Fischer~Calderon, M.~Scharff, J.~H. Alencastre~Miranda, and
  C.~Behn, ``Effects of multi-point contacts during object contour scanning
  using a biologically-inspired tactile sensor,'' \emph{Sensors}, vol.~20,
  no.~7, p. 2077, 2020.

\bibitem{9830882}
Z.~Yu, Y.~Guo, J.~Su, Q.~Huang, T.~Fukuda, C.~Cao, and Q.~Shi, ``Bioinspired,
  multifunctional, active whisker sensors for tactile sensing of mobile
  robots,'' \emph{IEEE Robotics and Automation Letters}, vol.~7, no.~4, pp.
  9565--9572, 2022.

\bibitem{9982122}
M.~A. Lin, E.~Reyes, J.~Bohg, and M.~R. Cutkosky, ``Whisker-inspired tactile
  sensing for contact localization on robot manipulators,'' in \emph{2022
  IEEE/RSJ International Conference on Intelligent Robots and Systems (IROS)},
  2022, pp. 7817--7824.

\bibitem{8968518}
S.~Kim, C.~Velez, D.~K. Patel, and S.~Bergbreiter, ``A magnetically transduced
  whisker for angular displacement and moment sensing,'' in \emph{2019 IEEE/RSJ
  International Conference on Intelligent Robots and Systems (IROS)}, 2019, pp.
  665--671.

\bibitem{9813357}
C.~Xiao, S.~Xu, W.~Wu, and J.~Wachs, ``Active multiobject exploration and
  recognition via tactile whiskers,'' \emph{IEEE Transactions on Robotics},
  vol.~38, no.~6, pp. 3479--3497, 2022.

\bibitem{10610850}
C.~Ye, G.~De~Croon, and S.~Hamaza, ``A biomorphic whisker sensor for aerial
  tactile applications,'' in \emph{2024 IEEE International Conference on
  Robotics and Automation (ICRA)}, 2024, pp. 5257--5263.

\bibitem{huet2017tactile}
L.~A. Huet, J.~W. Rudnicki, and M.~J. Hartmann, ``Tactile sensing with whiskers
  of various shapes: Determining the three-dimensional location of object
  contact based on mechanical signals at the whisker base,'' \emph{Soft
  robotics}, vol.~4, no.~2, pp. 88--102, 2017.

\bibitem{collinson2021tapered}
D.~W. Collinson, H.~M. Emnett, J.~Ning, M.~J. Hartmann, and L.~C. Brinson,
  ``Tapered polymer whiskers to enable three-dimensional tactile feature
  extraction,'' \emph{Soft Robotics}, vol.~8, no.~1, pp. 44--58, 2021.

\bibitem{9366394}
T.~A. Kent, S.~Kim, G.~Kornilowicz, W.~Yuan, M.~J.~Z. Hartmann, and
  S.~Bergbreiter, ``Whisksight: A reconfigurable, vision-based, optical whisker
  sensing array for simultaneous contact, airflow, and inertia stimulus
  detection,'' \emph{IEEE Robotics and Automation Letters}, vol.~6, no.~2, pp.
  3357--3364, 2021.

\bibitem{10160408}
T.~A. Kent, H.~Emnett, M.~Babaei, M.~J.~Z. Hartmann, and S.~Bergbreiter,
  ``Identifying contact distance uncertainty in whisker sensing with tapered,
  flexible whiskers,'' in \emph{2023 IEEE International Conference on Robotics
  and Automation (ICRA)}, 2023, pp. 607--613.

\bibitem{merker2021vibrissa}
L.~Merker, J.~Steigenberger, R.~Marangoni, and C.~Behn, ``A vibrissa-inspired
  highly flexible tactile sensor: scanning 3d object surfaces providing tactile
  images,'' \emph{Sensors}, vol.~21, no.~5, p. 1572, 2021.

\bibitem{10610224}
D.~Brouwer, J.~Citron, H.~Choi, M.~Lepert, M.~Lin, J.~Bohg, and M.~Cutkosky,
  ``Tactile-informed action primitives mitigate jamming in dense clutter,'' in
  \emph{2024 IEEE International Conference on Robotics and Automation (ICRA)},
  2024, pp. 7991--7997.

\bibitem{10473086}
S.~Lee, J.~I. Kim, Y.~Baek, D.~Chang, J.~Lee, Y.~S. Park, D.~Lee, and Y.-L.
  Park, ``Fiber-optic force sensing of modular robotic skin for remote and
  autonomous robot control,'' \emph{IEEE Transactions on Robotics}, vol.~40,
  pp. 2373--2389, 2024.

\bibitem{9562060}
S.~Suresh, M.~Bauza, K.-T. Yu, J.~G. Mangelson, A.~Rodriguez, and M.~Kaess,
  ``Tactile slam: Real-time inference of shape and pose from planar pushing,''
  in \emph{2021 IEEE International Conference on Robotics and Automation
  (ICRA)}, 2021, pp. 11\,322--11\,328.

\bibitem{10610762}
T.~Kossas, W.~Remmas, R.~Gkliva, A.~Ristolainen, and M.~Kruusmaa,
  ``Whisker-based tactile navigation algorithm for underground robots,'' in
  \emph{2024 IEEE International Conference on Robotics and Automation (ICRA)},
  2024, pp. 13\,164--13\,170.

\bibitem{8481577}
A.~Daniyan, S.~Lambotharan, A.~Deligiannis, Y.~Gong, and W.-H. Chen, ``Bayesian
  multiple extended target tracking using labeled random finite sets and
  splines,'' \emph{IEEE Transactions on Signal Processing}, vol.~66, no.~22,
  pp. 6076--6091, 2018.

\end{thebibliography}
\balance

\end{document}